\documentclass[11pt,a4paper]{article}

\usepackage[hyperref]{acl2019}
\usepackage{times}
\usepackage{latexsym}
\usepackage{amsmath}
\usepackage{amsfonts}
\DeclareMathOperator*{\argmax}{arg\,max}

\usepackage{caption}
\usepackage{multirow, makecell}
\usepackage{booktabs}
\usepackage{graphicx}

\aclfinalcopy 

\title{
Importance of Search and Evaluation Strategies\\ in Neural Dialogue Modeling
}

\author{Ilia Kulikov \\
  New York University \\
  {\tt kulikov@cs.nyu.edu} \\
  \And
  Alexander H. Miller \\
  Facebook AI Research \\
  \And
    Kyunghyun Cho \\
  New York University \\
  Facebook AI Research \\
  CIFAR Azrieli Global Scholar \\
  \And
    Jason Weston \\
  Facebook AI Research \\
  New York University \\
    }
\date{}

\begin{document}
\maketitle
\begin{abstract}
We investigate the impact of search strategies in neural dialogue modeling.
We first compare two standard search algorithms, greedy and beam search, as well as our newly proposed iterative beam search which produces a more diverse set of candidate responses.
We evaluate these strategies in realistic full conversations with humans and propose a model-based Bayesian calibration to address annotator bias.
These conversations are analyzed using two automatic metrics: log-probabilities assigned by the model and utterance diversity. Our experiments reveal that better search algorithms lead to higher rated conversations. However, finding the optimal selection mechanism 
to choose from a more diverse set of candidates is still an open question.

\end{abstract}

\section{Introduction}

There are three high-level steps to building a neural autoregressive sequence model for dialog modeling, 
inspired by 
work of \citet{vinyals2015neural}. First, decide on a network architecture which will consume previous utterances as well as any extra information such as speaker identifiers. Second, choose a learning strategy. Finally, decide on a search algorithm, as neural autoregressive sequence models do not admit a tractable, exact approach for generating the most likely response.

Recent research in neural dialogue modeling has often focused on the first two aspects. A number of variants of sequence-to-sequence models~\citep{sutskever2014sequence,cho2014learning,kalchbrenner2013recurrent} have been proposed for dialogue modeling in recent years, including hierarchical models~\citep{serban2016building} and transformers~\citep{mazare2018training,yang2018learning}. These advances in network architectures have often been accompanied by advanced learning algorithms. \citet{serban2017hierarchical} introduce latent variables to their earlier hierarchical model and train it to maximize the variational lower bound, similar to \citet{zhao2017learning} who propose to build a neural dialogue model as a conditional variational autoencoder. \citet{xu2017} and \citet{li2017adversarial} train a neural dialogue model as conditional generative adversarial networks~\citep{mirza2014conditional}.  These two learning algorithms, variational lower-bound maximization and adversarial learning, have been combined into a single model by \citet{shen2018improving}, which has been followed by \citet{gu2018dialogwae}. 

Despite abundant endeavors on modeling and learning, search has received only a little attention \citep{dinan2019second}. Most of the work on search has focused on training an additional neural network that provides a supplementary score to guide either greedy or beam search. \citet{li2015diversity} propose a maximum mutual information criterion for decoding using a reverse model. This has been extended by \citet{li2017learning}, where an extra neural network is trained to predict an arbitrary reward given a partial hypothesis and used during decoding. Similarly, \citet{zemlyanskiy2018aiming} train a neural network that predicts the other participant's personality given a partial conversation and use its predictability as an auxiliary score for re-ranking a set of candidate responses. None of these approaches study how the choice of the underlying search algorithm, rather than its scoring function, affects the quality of the neural dialogue model.

In this paper, we investigate the effects of varying search and selection strategies on the quality of generated dialogue utterances.
We start with 
an attention-based sequence-to-sequence model~\citep{bahdanau2014neural} trained on the recently-released PersonaChat dataset~\citep{zhang2018personalizing}. 
We evaluate three search algorithms: greedy search, beam search and iterative beam search, the last of which we design based on earlier works by \citet{batra2012diverse}. These algorithms are qualitatively different from each other in the size of subspace over which they search for the best response.

We compare all of these alternatives using two families of metrics. First, we use human evaluation of full, multi-turn conversation. The resulting distribution of annotator's scores has huge variance that is rarely discussed nor analyzed by other groups. This variance comes from each annotator's individual attitude towards and understanding of the task, which we call annotator bias.
In order to address this bias, we propose model-based Bayesian calibration that explicitly factors in each annotator's bias and the algorithm's underlying score, and report the posterior mean and variance of each algorithm's score. Additionally, we also compare automatic metrics that capture the model's intrinsic preference (log-probability) and the diversity of responses (distinct-$n$).

We make two key observations from the experiments. A better search strategy can indeed generate
responses that are both intrinsically preferred by the underlying model and diverse, without re-designing or re-training the neural dialogue model.
However, this observation does not necessarily carry over to human evaluation, as the best performing strategy according to these automatic metrics was not the best strategy according to human annotators. These results highlight both the importance of search algorithms as well as the difficulty in evaluating neural dialogue systems in a realistic, full conversation setup. 

We will make  trained models, code and human evaluation transcripts publicly available. Randomly sampled transcripts for each strategy are available in 2 additional pages of examples. All transcripts are given in additional materials and we encourage everyone to read it.

\section{Neural dialogue modeling}

Since \citet{vinyals2015neural}, a neural autoregressive sequence model based on sequence-to-sequence models~\citet{sutskever2014sequence,cho2014learning} have become one of the most widely studied approaches to dialogue modeling~\citep[see, e.g.,][]{serban2016building,serban2017hierarchical,zhao2017learning,xu2017,li2016simple,li2017learning,li2017adversarial,zemlyanskiy2018aiming,zhang2018personalizing,miller2017parlai,shen2018improving,gu2018dialogwae}. In this approach, a neural sequence model is used to model a conditional distribution over responses given a context which consists of previous utterances by both itself and a partner in the conversation as well as any other information about the speaker. 

\subsection{Neural autoregressive sequence modeling}
\label{sec:seq2seq}

A neural autoregressive sequence model learns the conditional distribution over all possible responses given the context.
Each conditional distribution is modelled by a neural network, and popular choices include recurrent neural networks~\citep{mikolov2010recurrent,sutskever2014sequence,cho2014learning,bahdanau2014neural}, convolutional networks~\citep{dauphin2016language,gehring2017convolutional} and self-attention~\citep{sukhbaatar2015end,vaswani2017attention}.
We explore search strategies and fix the model to a recurrent neural network.

\paragraph{Learning: Maximum Likelihood}

Each example in a training set $D$ consists of auxiliary information or context $U$ (such as a persona profile or external knowledge context) and a sequence of utterances, each of which is marked with a speaker tag, i.e.,
$C = (U, (Y_1^a, Y_1^b, \ldots, Y_L^a, Y_L^b) \in D$,
where $Y_l^s$ is the utterance from the $l$-th turn by a speaker $s$. The conditional log-probability assigned to this example given by a neural sequence model is then written as
\begin{align}
\label{eq:logp_conv}
    \log p(C) = \sum_{s \in \left\{ a, b \right\}} 
    \sum_{l=1}^L \log p(Y_l^s|Y_{<l}^s, Y_{\leq l}^{\bar{s}}, U),
\end{align}
where $\bar{s} = a$ if $s=b$ and otherwise $\bar{s}=b$.

Learning maximizes the log-probabilities of all the conversations in the training set:
\begin{align}
    \label{eq:loglikelihood}
    L = \frac{1}{|D|} \sum_{C \in D} \log p(C),
\end{align}
often done using stochastic gradient descent with backpropagation~\citep{rumelhart1985learning}.

\subsection{Inference (generation)}
\label{sec:inference}

In this paper, we generate a response to the current state of the conversation (but do not attempt to plan ahead to future exchanges), maximizing 
\begin{align*}
\log p(Y|Y_{<l}^s, Y_{<l}^{\bar{s}}, U) =
\sum_{t=1}^T \log p(y_t | y_{<t}, Y_{<l}^s, Y_{<l}^{\bar{s}}, U).
\end{align*}

Unfortunately, it is intractable to solve this problem due to the exponentially-growing space of all possible responses w.r.t. the maximum length $T$. It is thus necessary to resort to approximate search algorithms. 

\paragraph{Greedy search}

Greedy search has been the search algorithm of choice among the recent papers on neural dialogue modeling~\citep{gu2018dialogwae,zhao2017learning,xu2017,weston2018retrieve,zhang2018personalizing}.
It moves from left to right selecting one token at a time, simply choosing the most likely token at the current time step:
\begin{align*}
    \hat{y}_t = \argmax_{v \in V} \log p(y_t=v|\hat{y}_{<t}, Y_{<l}^s, Y_{<l}^{\bar{s}}, U).
\end{align*}

Greedy search has been found significantly sub-optimal within the field of machine translation ~\citep[see, e.g., Table~1 in][]{chen2018stable}, where similar neural sequence models are frequently used.

\paragraph{Beam search}
\label{sec:beamsearch}

Instead of maintaining a single hypothesis at a time, as in greedy search above, at time step $t$ beam search maintains a set of $K$ hypotheses $\mathcal{H}_t$:
\begin{align}
\label{eq:beam_partial}
    \mathcal{H}_t &= \{ (y_1^1, \dots, y_t^1), \dots, (y_1^K, \dots, y_t^K) \}. 
\end{align}
Each hypothesis $h^i_{y^i_t}, i \in \{1,\dots, K\}$ from $\mathcal{H}_t$ is expanded with all possible next tokens $v$ from the vocabulary $V$ to form candidate hypotheses. Each candidate is in the form of
\begin{align}
\label{eq:candidate_hyp}
    \tilde{h}^i_v = h^i_{y_t}\|(v) = (y_1^i, \ldots, y_t^i, v),
\end{align}
and is assigned a score:
\begin{align}
    \label{eq:candidate_score}
    s(\tilde{h}^i_v) = s(h^i_{y_t^i}) + \log p(v|y_{\leq t}^i).
\end{align}

The new hypothesis set of $K$ hypotheses is then constructed as
\begin{align}
    \mathcal{H}_{t+1} = \underset{i,v}{\text{arg-top-}k}~~
    s(\tilde{h}^i_v).
    \label{eq:topk}
\end{align}

From the new hypothesis set, we find and copy \emph{finalized} hypotheses (sequences ending with the special token $\left<\text{eos}\right>$ for ``end of sequence'') to a candidate sequence set $\mathcal{M}_t$. That is,
\begin{align*}
    \mathcal{M}_t = \left\{ 
    h^i_v \in \mathcal{H}_{t+1} | v = \left<\text{eos}\right>
    \right\}.
\end{align*}

Beam search terminates when $| \cup_{t'=1}^t \mathcal{M}_t | \geq K'$, where $K'$ is the maximum number of candidate sequences to be returned, or when $t \geq L_{\max}$, where $L_{\max}$ is the maximum length allowed for each candidate sequence. When terminated, beam search returns all the candidate sequences in $\mathcal{M}=\cup_{t'=1}^t \mathcal{M}_t$. 

One can increase the size of the subspace over which beam search searches for a response and size of $\mathcal{M}$ by changing hyper-parameters $K,K',L_{\text{max}}$.
However, beam search is known to suffer from the problem that most of the hypotheses discovered in $\mathcal{M}$ are near each other in the response space \cite{li2016simple, li2015diversity}.
For tasks such as dialogue modeling, which are much more open-ended than e.g. machine translation, this is particularly troublesome as many high quality responses may be missing in the beam. 

\paragraph{Final sequence selection}
\label{par:selection}

We consider search strategies to produce a set of candidate responses for the model to choose from. While greedy search provides only a single possible sequence, beam search generates a candidate set of size $|\mathcal{M}|$. It is usual practice to use the score $s(h)$ used during the search to select the final sequence, but it is an open question whether there are better selection strategies for choosing between these final candidate responses.

\paragraph{Avoiding repeating $n$-grams}

Although this has not been reported in a formal publication in the context of neural dialogue modeling to our knowledge, \citet{paulus2017deep} and \citet{opennmt}
implement so-called $n$-gram blocking. In $n$-gram blocking, a hypothesis in a beam $\mathcal{H}_t$ is discarded if there is an $n$-gram that appears more than once within it.

\section{Uncovering hidden responses}

We now propose an improved search strategy. 
To address the locality 
issue in beam search, we propose an iterative beam search to radically increase the size of search space without introducing much computational overhead, inspired by earlier work on diverse beam search~\citep{vijayakumar2018diverse,batra2012diverse,li2016simple}. 

\subsection{Iterative beam search}
\label{sec:ibs}

The search space over which beam search has operated can be characterized by the union of all partial hypothesis sets $\mathcal{H}_t$ in Eq.~\eqref{eq:beam_partial}:
$\mathcal{S}_{0} = \cup_{t=1}^T \mathcal{H}_t,$
where we use the subscript $0$ to indicate that beam search has been done without any other constraint. Re-running beam search with an increased beam width $K$ would result in the search space that overlaps significantly with $\mathcal{S}_0$, and would not give us much of a benefit with respect to the increase in computation. 

Instead, we keep the beam size $K$ constant but run multiple iterations of beam search while ensuring that any previously explored space
\mbox{$\bar{\mathcal{S}}_{<l} = \cup_{l'=0}^{l-1} \mathcal{S}_{l'}$}
is {\it not} included in a subsequent iteration of beam search. This is done by setting the score of each candidate hypothesis $s(\tilde{h}_{t+1}^i)$ in Eq.~\eqref{eq:candidate_score} to  negative infinity, when this candidate is included in $\bar{\mathcal{S}}_{<l}$. 
We relax this inclusion criterion by using a non-binary dissimilarity metric, and say that the candidate is included in $\bar{\mathcal{S}}_{<l}$, if\begin{align}
\label{eq:ibeam-criterion}
    \min_{h \in \bar{\mathcal{S}}_{<l}}
    \Delta(\tilde{h}_{t+1}^i, h) < \epsilon,
\end{align}
where $\Delta$ is a string dissimilarity measure, such as Hamming distance used in this work, and $\epsilon$ is a similarity threshold. 

This procedure ensures that a new partial hypothesis set of beam search in the $l$-th iteration minimally overlaps with any part of the search space explored earlier during the first $l-1$ iterations of beam search.
By running this iteration multiple times, we end up with a set of top hypotheses from each iteration of beam search, from which the best one is selected according to for instance the log-probability assigned by the model.
We build a final candidate set $\mathcal{M}$ as a set of all these best hypotheses from beam search iterations.

\paragraph{Practical implementation}

A major issue with iterative beam search in its naive form is that it requires running beam search multiple times, when even a single run of beam search can be prohibitively slow in an interactive environment, such as in dialogue generation.
We address this computational issue by performing these many iterations of beam search in parallel simultaneously. At each time step in the search, we create sets of candidate hypotheses for all iterations in parallel, and go through these candidate sets in sequence from the $(l=0)$-th iteration down to the last iteration, while eliminating those candidates that satisfy the criterion in Eq.~\eqref{eq:ibeam-criterion}. We justify 
this parallelized approach by defining the similarity measure $\Delta$ to be always larger than the threshold $\epsilon$ when the previous hypothesis $h$ is longer than  $\tilde{h}^i_{t+1}$ in Eq.~\eqref{eq:ibeam-criterion}.

\section{Dialogue evaluation}

Broadly there are two ways to evaluate a neural dialogue model. The first approach is to use a set of (often human generated) reference responses and compare a single generated response against them \cite{serban2015survey,liu2016not}. There are several methods for this comparison: (1) measure the perplexity of reference responses using the neural dialogue model, (2) compute a string match-based metric of a generated response against reference responses, and (3) use human annotators to compare model generated responses against reference or other models' responses.  None of these approaches capture the effectiveness of a neural sequence model in conducting a full conversation, 
because the model responses are computed given a human-written context, i.e., it does not see its own responses in the dialogue history, but gold responses only.

We concentrate on a second approach for evaluation where a neural dialogue model has a multi-turn conversation with a human partner (or annotator) \cite{zhang2018personalizing, zemlyanskiy2018aiming, weston2018retrieve, dinan2019second}. Unlike other approaches, it requires active human interaction, as a conversation almost always deviates from a previously collected data even with the same auxiliary information ($U$ in Eq.~\eqref{eq:logp_conv}).
This evaluation strategy reflects both how well a neural dialogue model generates a response given a correct context as well as how well it adapts to a dynamic conversation---the latter was not measured by the first strategy, where the model only had to generate a single response.

\subsection{Human evaluation of a full conversation}
\label{sec:humeval}

An annotator is asked to make a conversation with a randomly selected model (search strategy) for at least five turns.
At the end of the conversation, we ask the annotator three sets of questions:
\begin{enumerate}
\itemsep -0.4em
    \item Overall score ($\left\{ 1, 2, 3, 4\right\}$)
    \item Marking of each {\it good} utterance-pair ($\left\{ 0, 1 \right\}$)
    \item Marking of each {\it bad} utterance-pair ($\left\{ 0, 1 \right\}$)
\end{enumerate}

The first overall score allows us to draw a conclusion on which algorithm makes a better conversation overall. We use a 4 point scale in order to avoid having a ``catch-all'' category in the answer \citep{dalal2014middle}.
The latter two questions are collected to investigate the relationship between the overall impression and the quality of each utterance-pair. 

\subsection{Bayesian calibration}
\label{sec:bayesian}

Although human evaluation is desirable, 
raw scores collected by annotators are difficult to use directly due to the annotator bias.
Some are more generous while others are quite harsh,
as recently reported in 
\citet{zhang2018personalizing,zemlyanskiy2018aiming}.
We propose using Bayesian inference as a framework to account for the bias of each annotator, and describe two instances of this framework below.

\paragraph{1-4 star rating of a conversation} 

We treat both the unobserved score $M_i$ of each model,
in our case each search algorithm, and the unobserved bias $B_j$ of each annotator as latent variables. The score of the $i$-th model follows the following distribution:
    $\mu_i \sim \mathcal{U}(1, 4)\text{ and }
    M_i \sim \mathcal{N}(\mu_i, 1^2)$, 
where $\mathcal{U}$ and $\mathcal{N}$ are uniform and normal distributions. It states that {\it a priori} each model is likely to be uniformly good or bad. 
The annotator bias $B_j$ follows
    \mbox{$B_j \sim \mathcal{N}(0, 1^2)$},
where we are assuming that each annotator does not have any bias {\it a priori}. 

\begin{figure}[t]
\centering
\includegraphics[width=\columnwidth]{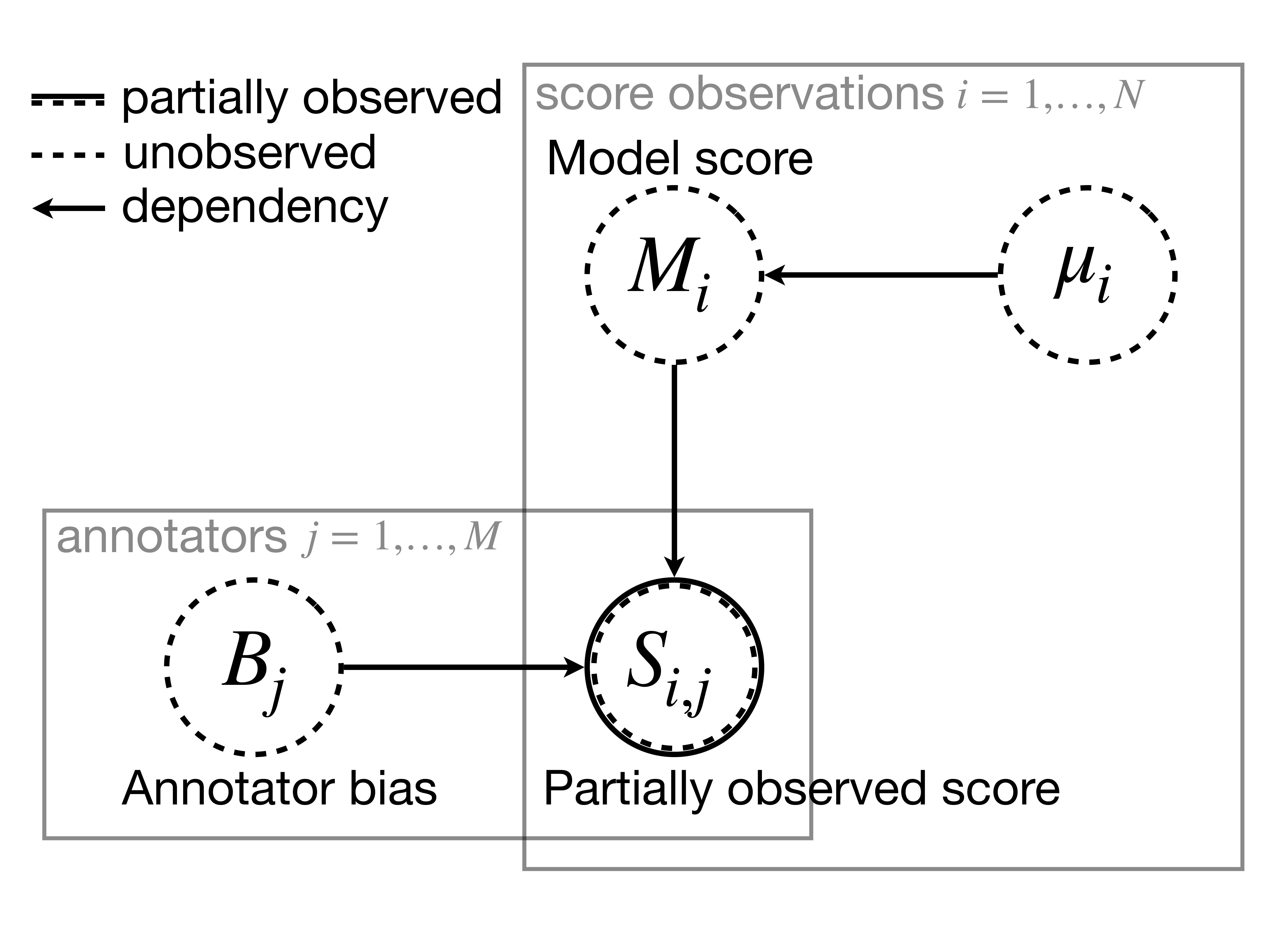}
\caption{Graphical model used for Bayesian Calibration. $M$ annotators participated such that in total $N$ observed scores are presented.}
\label{fig:bayesianmodel}
\end{figure}

Given the model score $M_i$ and annotator bias $B_j$, the conditional distribution over an observed score $S_{ij}$ given by the $j$-th annotator to the $i$-th model is then:
\begin{align*}
    S_{ij} \sim \mathcal{N}(M_i+B_j, 1^2).
\end{align*}
Due to the nature of human evaluation, only a few of $S_{ij}$'s are observed. 
Figure~\ref{fig:bayesianmodel} shows the graphical model described above.

The goal of inference in this case is to infer the posterior mean and the variance: 
\begin{align}
&\mathbb{E}[M_i | \left\{ S_{ij} | S_{ij} \in \mathcal{O} \right\}],
\label{eq:posterior_mean}
\\
&\mathbb{V}[M_i | \left\{ S_{ij} | S_{ij} \in \mathcal{O} \right\}],
\nonumber
\end{align}
where $\mathcal{O}$ is a set of observed scores. 

\paragraph{Binary rating of an utterance}

When an annotator labels pairs of utterances from the conversation with a binary score $\left\{ 0, 1 \right\}$ (such as whether that pair was a ``good'' exchange), we need to further take into account the turn bias $T_k$:
    \mbox{$T_k \sim \mathcal{N}(0, 1^2)$}.
As we will use a Bernoulli distribution for each observed score rather than a 1-4 rating, we modify the prior of the model scores accordingly: \mbox{$M_i \sim \mathcal{N}(0, 1^2).$}

The distribution of an observed utterance-pair score 
is then
    \mbox{$S_{ijk} \sim \mathcal{B}(\text{sigmoid}(M_i+B_j+T_k))$},
where $\mathcal{B}$ is a Bernoulli distribution.
The goal of inference is then to compute 
\begin{align}
\label{eq:posterior_mean_binary}
    \mathbb{E}_{M_i | \left\{ S_{ijk} | S_{ijk} \in \mathcal{O} \right\}} \left[
    \text{sigmoid}(M_i)
    \right], 
    \\
    \nonumber
        \mathbb{V}_{M_i | \left\{ S_{ijk} | S_{ijk} \in \mathcal{O} \right\}} \left[
    \text{sigmoid}(M_i)
    \right],
\end{align}
which estimate the average number of positively labelled utterance-pairs given the $i$-th model and the uncertainty in this estimate, respectively.

\paragraph{Inference}

We use no-u-turn (NUTS) sampler~\citep{hoffman2014no} for posterior inference in Pyro~\citep{bingham2018pyro}.

\section{Experiment Settings}

\subsection{Data: Persona-Chat}

We use Persona-Chat, released recently by \citet{zhang2018personalizing} and the main dataset for the Conversational Intelligence Challenge 2 (ConvAI2),\footnote{
\url{http://convai.io/}
}
to train a neural dialogue model.
The dataset contains dialogues between pairs of speakers randomly assigned personas from a set of 1,155, each consisting of 4-5 lines of description about the part they should play, e.g. {\em ``I have two dogs''} or  {\em``I like taking trips to Mexico''}.
The training set consists of 9,907 such dialogues where pairs of partners play their roles, and a validation set of 1,000 dialogues. The ConvAI2 test set has not been released. Each dialogue is tokenized into words, resulting in a vocabulary of 19,262 unique tokens. See  \citet{zhang2018personalizing} for more details.

\subsection{Neural dialogue modeling}

\paragraph{Model}

We closely follow \citet{bahdanau2014neural} in building an attention-based neural autoregressive sequence model. The encoder has two bidirectional layers of 512 LSTM \citep{hochreiter1997long} units each direction-layer, and the decoder has two layers of 512 LSTM units each. We use global general attention as described by \citet{luong2015effective}. We use the same word embedding matrix for both the encoder and decoder, which is initialized from 300-dimensional pretrained GloVe vectors~\citep{pennington2014glove} for the 97\% of the vocabulary which overlaps with GloVe. We allow word embedding weights to be updated during the training. 

\paragraph{Learning}

We use Adam~\citep{kingma2014adam} with the initial learning rate set to $0.001$. We apply dropout~\citep{srivastava2014dropout} between the LSTM layers with the dropout rate of $0.5$ to prevent overfitting. We train the neural dialogue model until it early-stops on the validation set.\footnote{
When the validation loss 
does not improve for twelve epochs, we early-stop.
} 

The perplexity of trained model on the ConvAI2 validation set is $24.84$, which is competitive compared to the other entries on the competition's leaderboard.\footnote{
\url{https://github.com/DeepPavlov/convai/blob/master/leaderboards.md}}
Our model serves well as an underlying system for investigating the effect of search algorithms.

\subsection{Search Strategies}

We test three search strategies; {\bf greedy} and {\bf beam} search from \S\ref{sec:inference}, and iterative beam search ({\bf iter-beam}) from \S\ref{sec:ibs}.

Beam search ({\bf beam}) uses beam size $K = 5$ and $K' = 15$.
This decision is based on preliminary experiments where we found that smaller beam sizes work better than larger ones do. We use the length penalty, described by \citet{wu2016google} and $n$-gram blocking from \S\ref{sec:inference}.

Iterative beam search ({\bf iter-beam}) uses $15$ iterations of beam search with beam size $5$ resulting in a candidate set of size $15$. We use the same length penalty and $n$-gram blocking as in beam search ({\bf beam}).
Given the hyper-parameters above both {\bf beam} and {\bf iter-beam} produce $15$ candidates and selects the final response based on log-probability.

\begin{figure}[t]
\centering
\includegraphics[width=\columnwidth]{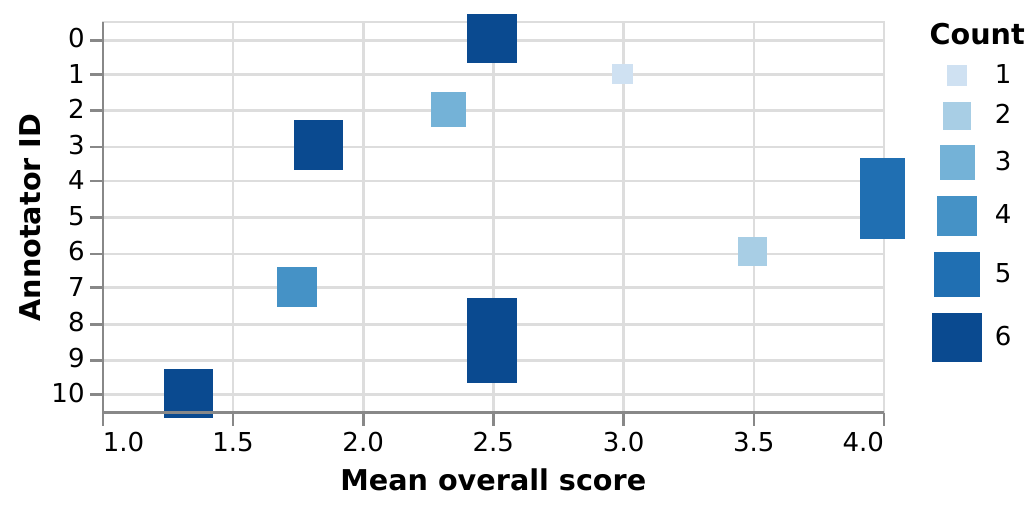}
\caption{
\label{fig:annotator-bias}
The distribution of averaged overall scores given by annotators to greedy search ({\bf greedy}). Each row plots scores given by a single annotator over multiple conversations. Counts show how many dialogues each annotator performed.
}

\vspace{-6mm}
\end{figure}

\begin{table*}[t!]
\centering
\begin{tabular}{cccccccc}
      Search & log-p$\uparrow$ & \multicolumn{2}{c}{Overall Score (1-4)$\uparrow$}    & \multicolumn{2}{c}{\% Good Pairs$\uparrow$}   & \multicolumn{2}{c}{\% Bad Pairs$\downarrow$}  \\
strategy & &Raw&Calibrated& Raw&Calibrated&Raw&Calibrated \\
\toprule
greedy          &-9.66$\pm$2.73 & 2.56$\pm$0.98     & 2.30$\pm$0.24 & 0.45 & 0.28$\pm$0.07 & 0.38 & 0.54$\pm$0.07  \\
beam            &-7.26$\pm$2.28& 2.67$\pm$0.86     & 2.70$\pm$0.27 & 0.58 & 0.44$\pm$0.08 & 0.35 & 0.27$\pm$0.01  \\
iter-beam       &{\bf -5.95$\pm$1.35} & 2.80$\pm$0.90     & 2.67$\pm$0.23 & 0.58 &  0.45$\pm$0.08 & 0.31 & 0.32$\pm$0.03  \\
\midrule
human &-42.95$\pm$18.87& 3.62$\pm$0.71 & 3.37$\pm$0.22 & 0.76 & 0.76$\pm$0.06 & 0.07 & 0.04$\pm$0.003 \\
\end{tabular}

\caption{ The average log-probabilities and model scores (average$\pm$standard deviation) assigned to the responses during human evaluation. Better search algorithms find responses with higher log-probabilities according to the model.
Without observing standard deviations and calibrated scores one can make erroneous conclusions. 
}

\label{tab:human-eval}
\end{table*}

\subsection{Evaluation}

\paragraph{Human evaluation}

We use ParlAI~\citep{miller2017parlai} which provides seamless integration with Amazon Mechanical Turk (MTurk) for human evaluation.
A human annotator is paired with a model with a specific search strategy, and both are randomly assigned personas out of a set of 1,155, and are asked to make a conversation of at least either five or six turns (randomly decided). We allow each annotator to participate in at most six conversations per search strategy and collect approximately 50 conversations per search strategy and additional human-human test.\footnote{
Some conversations were dropped due to technical errors,
resulting in total 50, 51, 49 and 53 conversations for {\bf greedy}, {\bf beam}, {\bf iter-beam} and {\bf humans}, respectively. 
} 
Each conversation is given a single overall score and two sequences of binary utterance pairs flags, as described in \S\ref{sec:humeval}. 

\paragraph{Bayesian calibration}

In order to remove annotator bias, or inter-annotator variability, we use Bayesian calibration from \S\ref{sec:bayesian}. We take 50 warm-up steps and collect 150 samples using NUTS sampler for inferring the posterior mean and variance of the model score in Eq.~\eqref{eq:posterior_mean}. We use 30 warm-up steps and 100 samples for inferring the mean and variance of the average portion of positively or negatively labelled utterance-pairs in Eq.~\eqref{eq:posterior_mean_binary}.\footnote{Variances of inferred posterior distribution and original data distribution are not comparable, as the former reflects the uncertainty in posterior inference rather than the spread of scores.}

\paragraph{Automatic metrics}

In addition to human evaluation, we compute automatic metrics to quantitatively characterize each search algorithm. First, we report the \mbox{{\bf log-p}robability} of a generated response assigned by the model which is a direct indicator of the quality of a search algorithm. Second, we compute the average number of unique $n$-grams generated per conversation normalized by the number of generated tokens in the conversation, called {\bf distinct-$n$} from \citep{li2015diversity}, with $n=1,2,3$.  

We compute distinct-$n$ in two different settings. First, we compute distinct-$n$ over the candidate set $\mathcal{M}$ given by the search algorithm. Second, we compute distinct-$n$ over the final selected responses for each search strategy. The former shows diversity within the possible response candidates, while  the  latter  shows  diversity  among  the  actual selected dialogue  outputs.

\section{Result}
\label{sec:results}

\subsection{Human Evaluation}

\paragraph{Annotator bias}

In Fig.~\ref{fig:annotator-bias}, we plot the averaged scores provided by the human annotators for one search strategy ({\bf greedy}), where each row corresponds to each annotator. 
Consider the three annotators with id $3,4,10$. Their means are clearly separated from each other, which points to the existence of annotator bias. 
This observation  supports the necessity of the Bayesian calibration described in \S\ref{sec:bayesian}.

\paragraph{Human evaluation}
\label{sec:humanresults}

In Table~\ref{tab:human-eval}, we present the scores from human evaluation. In total 41 unique annotator participated within 201 collected conversations. 
We make a major observation which is that greedy search ({\bf greedy}), which has been the search algorithm of choice in neural dialogue modeling, significantly lags behind the variants of beam search ({\bf beam, iter-beam}) in all metrics. This stark difference is worth our attention, as this difference is {\it solely} due to the choice of a search algorithm and is not the result of different network architectures nor learning algorithms. In fact, this cannot even be attributed to different parameter initialization, as we use only {\it one} trained model for all of these results.

The model scores assigned to human conversations ({\bf humans}) are far superior to all search strategies. It is clear with both overall score and utterance pairs proportion scores. This tells us that there are many open questions how to improve neural dialogue models.

\subsection{Automatic Metrics}

\paragraph{Search quality: log probability ({\bf log-p})}

Better search algorithms find responses with higher log-probability according to the model, as shown in Table~\ref{tab:human-eval}. This is a natural consequence from exploring a larger subset of the search space. 

A notable observation from Table~\ref{tab:human-eval} is that the neural sequence model assigns very low log-probabilities to human response. This implies that there is a limit to improving this specific neural dialogue model by using a better search algorithm and instead there is more room to improve the models and learning algorithms to place a high probability on human responses. It is necessary to test the proposed search strategies with new models and we leave this for the future.

\begin{table*}[t!]
\centering
\begin{tabular}{ccccccc}
       & \multicolumn{6}{c}{distinct-$n$ $\uparrow$}  \\
        Search & \multicolumn{2}{c}{$n=1$} &  \multicolumn{2}{c}{$n=2$} &  \multicolumn{2}{c}{$n=3$} \\
       strategy & post & pre & post & pre & post & pre \\
\toprule
greedy      &    0.47 & - & 0.61 & - & 0.62 & - \\
beam        &    0.56 & 0.06 & {\bf 0.69} & 0.12 & {\bf 0.63} & 0.18 \\
iter-beam   &    {\bf 0.59} & {\bf 0.18} & 0.68 & {\bf 0.41} & 0.60 & {\bf 0.58} \\
\midrule
human & 0.66 & - & 0.85 & - & 0.82 & - \\
\end{tabular}

\caption{Measuring the diversity of different search and selection strategies. Both {\bf beam} and {\bf iter-beam} produce up to 15 hypotheses each. Column {\bf post} is distinct-$n$ measured over the final selected output responses given by the model and allows us to compare the diversity of the best responses each search procedure produces. Column {\bf pre} is distinct-$n$ measured over the candidate set $\mathcal{M}$ given by the search algorithm and allows us compare diversity generated within the search.
}
\label{tab:distinc}
\end{table*}

\begin{table*}[t!]
\centering
\scriptsize
\begin{tabular}{ll}
{\bf Beam search} & {\bf Iterative beam search} \\
\toprule
do you have any pets ? & that ' s cool , what do you like to eat ? \\
what is your favorite animal ? & do you have a favorite color ? mine is pink . \\
i like to talk about strangers . & what do you like to eat ? \\
do you like animals ? & i don ' t like fish , but my favorite color is pink . \\
do you like animals ? i want to live at the beach . & what color is your hair ? \\
do you like animals ? i ' ve a pet . & that does sound good , i like to go alone . \\
do you like animals ? i want to live on the beach . & i would love to eat fish . \\
do you like animals ? i want to live at the beach & that makes sense , do you have any hobbies ? \\
do you like animals ? i ' ve a monkey . & i hear you , my favorite color is pink . \\
do you like animals ? i want to be a monkey . & i want to be a yoga instructor . \\
do you like animals ? i want to live at the beach , but love it . & i did not eat meat , but my favorite color is pink . \\
do you like animals ? i want to live at the beach , but love monkeys . & what are your favorite foods ? mine is pink . \\
do you like animals ? i want to live at the beach , but they are my favorite . & what does your favorite color ? mine is pink . \\
do you like animals ? i want to live at the beach , but have a monkey . & what type of food do you like ? \\
do you like animals ? i want to live at the beach , but they are my favorite & i spend a lot of time alone . \\
\end{tabular}
\caption{{\bf beam} and {\bf iter-beam} candidate sets $\mathcal{M}$. These are from one turn of one randomly selected conversation from human evaluation. {\bf iter-beam} produces more diverse responses.}
\label{tab:candidates}
\end{table*}

\paragraph{Diversity: distinct-$n$} The diversity metric is measured before (pre) and after (post) selecting the final response from the candidate set $\mathcal{M}$ for both {\bf beam} search and {\bf iter-beam} search. Since {\bf greedy} and {\bf humans} produce only single response, we compute the diversity metric only using those final responses for both greedy search and humans. In both pre and post settings, the normalization is done per each conversation.

As well as in human evaluation, {\bf greedy} has lower diversity compared to all the other strategies as shown in Table~\ref{tab:distinc}. 
We see large gap in pre-selection distinct-$n$ for all $n$ between {\bf beam} and {\bf iter-beam} while the difference is small in post-selection distinct-$n$. In other words, while providing more diverse set of candidates, the final selected output response with {\bf iter-beam} is not particularly diverse. This agrees well with human evaluation, where both {\bf iter-beam} and {\bf beam} model scores were indistinguishable, as annotators could only see the final response after selecting from the candidate set. Table~\ref{tab:candidates} shows pre-selection candidate sets for both beam search and iterative beam search.

Finally, we observe a significant gap between the best search strategy and humans in these diversity metrics. Together with the gap we observed in human evaluation scores, we suspect that the lack of diversity in the output responses is a major factor behind the low performance of the tested neural dialogue model in the human evaluation.

\section{Conclusion and Discussion}
We have performed realistic human evaluation of the neural dialogue model to validate the importance of exploring better search strategies. We observed that careful design of human evaluation is necessary to properly evaluate the ability of the neural dialogue model to conduct a conversation with a human. The proposed Bayesian calibration of model scores helps to account the annotator bias observed in human evaluation.

Extensive analysis reveals that greedy search, which has been the inference algorithm of choice in neural dialogue modeling, significantly lags behind more sophisticated search strategies such as beam search and iterative beam search.


We have proposed the iterative beam search which produces more diverse set of candidate responses w.r.t. pre-selection distinct-$n$ metric. Post-selection final responses with iterative beam search have higher log-probability compared to other search strategies. 
In spite of this, there is only a marginal difference between iterative beam search and beam search w.r.t. scores from human evaluation and post-selection distinct-$n$. This suggests that final response selection strategy is as important as the search strategy being used and can be a major factor in the inference pipeline of the neural dialogue model.
We leave improving this strategy to future work.

Finally, the model assigns lower probability to reference responses, which implies suboptimality in the current neural dialogue model . It is necessary in the future to test the proposed search strategies with new models.

\section*{Acknowledgments}
KC is partly supported by Samsung Advanced Institute of Technology (Next Generation Deep Learning: from Pattern Recognition to AI) and Samsung Electronics (Improving Deep Learning using Latent Structure).

\bibliography{acl2019}
\bibliographystyle{acl_natbib}

\newpage

\appendix

\begin{table*}[t]
\small
\centering
\begin{tabular}{l c r}
Annotator & & Model \\
\toprule
i lay awake at night so i can watch the stars. &
\multirow{4}{*}{\rotatebox{90}{Persona}}
& i like to hike and spend time outdoors. \\
i m often late for class. & & i am close to my mother , but not to my father. \\
i wish i was taller so i could reach the cookie jar. & & my favorite thing to eat is a burrito. \\
i want a porsche. & &  \\
i cuddle up with my dog every night. & & \\
\midrule
\fbox{Hello how are you?} & 0 & 
\fbox{I ' m doing well, how are you?} \\
\fbox{I'm great!  Waiting for dark so I can look at the stars!} & 1 & 
\fbox{\makecell[r]{that is cool, I love the outdoors, I love to hike.}} \\
\fbox{\makecell[l]{My dog and I love hiking, and cuddling!}}
&2  & \fbox{I love hiking, I love hiking, I love hiking.} \\
\fbox{\makecell[l]{
that's awesome!  I drive to the traiils in my porsche.}} &3  &
\fbox{I have a mustang, I ' ve a mustang, it is a nice car.} \\
\fbox{\makecell[l]{That's really cool!  I love sports cars.}} & 4 & \fbox{I like to go hiking, I love to eat burritos.} \\
\bottomrule
\end{tabular}
\caption{Search strategy: \textbf{greedy}. Annotator score: \textbf{3}}
\label{g2}
\end{table*}
    
\begin{table*}[t]
\small
\centering
\begin{tabular}{l c r}
Annotator & & Model \\
\toprule
    
 my father died when i was 2. & \multirow{4}{*}{\rotatebox{90}{Persona}} &  i believe in aliens. \\
 i have blue eyes. & &  \makecell[r]{to free my mind from feeling depressed \\ i look up at the stars at night.}\\
 i love fishing. & &  i cannot choose between lollipops and rainbows.\\
 my favorite season is spring. & &  i would love to be a professional eater and not get fat.\\
 & &  i hate cleaning up after my husband and children.\\
\midrule
\fbox{Hey there, how are you?} & 0& \fbox{I ' m good. how are you?} \\
\fbox{pretty good. I'm planning my next fishing trip!} & 1& \fbox{that sounds fun. I ' m going to be a professional eater} \\
\fbox{that sounds amazing! I would love that.} & 2& \fbox{do you have any hobbies?} \\
\fbox{I really like going fishing, especially in the spring time.} & 3& \fbox{I like to stay at night with my husband and kids} \\
\fbox{\makecell[l]{That is nice! I wish I had a family, my dad died \\ when I was really little.}} & 4& \fbox{I ' m sorry to hear that} \\

\bottomrule
\end{tabular}
\caption{Search strategy: \textbf{beam}. Annotator score: \textbf{4}}
\label{consistency}
\end{table*}

\begin{table*}[t]
\small
\centering
\begin{tabular}{l c r}
Annotator & & Model \\
\toprule
    
 i eat tuna fish salad at least every day. & \multirow{4}{*}{\rotatebox{90}{Persona}} &  i am a yoga instructor. \\
 my favorite music is two steps from hell and rock opera genre. & &  i married my highschool sweetheart.\\
 i have sandy brown hair and green eyes. & &  pink is my favorite color.\\
 i can t get enough gummy worms to satisfy my sweet tooth. & &  i enjoy being alone.\\
 i read sci fi space adventures with a passion. & & \\
\midrule
\fbox{hello there friend!} & 0& \fbox{hello, how are you?} \\
\fbox{I'm great. how are you?} & 1& \fbox{I ' m doing well, thank you.} \\
\fbox{What's your favorite food?} & 2& \fbox{I love pizza, what about you?} \\
\fbox{tuna fish salad is my favorite. i eat it every day.} & 3& \fbox{do you have a favorite color? mine is pink.} \\
\fbox{I like all of the colors, but my eyes are green. do you have any hobbies?} & 4& \fbox{I like to go alone.} \\
\fbox{ah, I enjoy reading sci-fi. that's something you do alone.} & 5& \fbox{do you have any hobbies?} \\
\fbox{reading. I also really like rock operas.} & 6& \fbox{what is your favorite genre?} \\
\fbox{rock opera. but my favorite band is two steps from hell. what about you?} & 7& \fbox{I like all kinds of music.} \\
\fbox{is there any food you can't get enough of?} & 8& \fbox{I do not have a lot of free time.} \\
\fbox{ah, sorry to hear that. It was nice chatting with you, though!} & 9& \fbox{it was nice talking to you too!} \\

\bottomrule
\end{tabular}
\caption{Search strategy: \textbf{iter-beam}. Annotator score: \textbf{3}}
\label{tab:finalex}
\end{table*}

\section{Selected dialogue transcripts from the evaluation}

We publish dialogue transcripts from human evaluation for reader's analysis. We randomly select a transcript per each search strategy. Formatted representations are printed in tables from Table~\ref{g2} to Table~\ref{tab:finalex}. The second row refers to personalized context from PersonaChat dataset \citep{zhang2018personalizing}. The third row prints the whole dialogue where each turn is bounded with a box. Left column named \texttt{Annotator} contains responses written by the annotator. Middle column prints positional ordering of turns. Right column named \texttt{Model} contains responses generated by the model. The caption contains the search type and the score given by the annotator. We have prepared all evaluation scripts for reader in additional materials and we encourage everyone to read it.

\end{document}


\maketitle

\appendix

\section{Human evaluation questionnaire}
\subsection{Overall scoring question}

Right after the end of dialogue system asks worker the following question:

Now the conversation is completed! Please evaluate the
conversation by clicking a button with score from [1, 2, 3, 4]
below, this score should reflect how you liked
this conversation (1 means you did not like it at all, and 4 means it was an engaging conversation).

\subsection{Good/bad pairs selection}

After the first question system asks following questions:

Now please select every interaction pair which you consider as a {\bf good}, natural pair of messages. Do not compare them between each other, try to use your life experience now.

Now please select every interaction pair which you consider as a {\bf bad}, some examples of bad partner response are: not answering your question, answering different question, random content, contradicts previous statements etc.

\section{Model/Training hyperparameters}
\begin{tabular}{ |l|r| }
\hline
RNN type & LSTM \\
RNN layers & 2 \\
hidden dim & 512 \\
embedding dim & 300 \\
dropout rate & 50\% \\
attention type & global general \\
bidirectional encoder & True \\
shared weights & enc/dec embeddings \\
\hline
negative samples & 5 \\
margin for ranking loss & 1.0 \\
fine-tuning rank weight & 1.0 \\
fine-tuning generation weight & 0.1 \\
\hline
batch size & 64 \\
optimizer & Adam \\
starting learning rate & 0.001 \\
gradient clip threshold & 0.1 \\
embedding pretraining & glove 840B \\
validation every... & 0.5 epochs \\
validation metric & hits@1 \\
max valid patience & 12 epochs \\
\hline
\end{tabular}